\documentclass[10pt,twocolumn,letterpaper]{article}

\usepackage{caption}
\usepackage{subcaption}
\usepackage{3dv}
\usepackage{times}
\usepackage{epsfig}
\usepackage{graphicx}
\usepackage{amsmath}
\usepackage{amssymb}
\usepackage[numbers,sort&compress]{natbib}
\usepackage{afterpage}
\usepackage{float}
\usepackage{wrapfig}

\newif\ifincludeappendix
\includeappendixtrue

\newif\ifappendixonly
\appendixonlyfalse

\newcommand{\norm}[1]{\left\lVert#1\right\rVert}

\newcommand{\figref}[1]{Figure~\ref{#1}}

\newcommand{\secref}[1]{Section~\ref{#1}}

\newcommand{\tabref}[1] {\reftab{#1}}
\newcommand{\reftab}[1] {Table~\ref{#1}}

\usepackage[pagebackref=true,breaklinks=true,letterpaper=true,colorlinks,bookmarks=false]{hyperref}
\threedvfinalcopy

\begin{document}
    
    \ifappendixonly
        \title{Convolutional Autoencoders for Human Motion Infilling\\Appendix}
    \else
        \title{Convolutional Autoencoders for Human Motion Infilling}
    \fi
    
    \author{Manuel Kaufmann\textsuperscript{1} \quad Emre Aksan\textsuperscript{1} \quad Jie Song\textsuperscript{1} \quad Fabrizio Pece\textsuperscript{2} \quad Remo Ziegler\textsuperscript{2} \quad Otmar Hilliges\textsuperscript{1} \\
    \\
    \textsuperscript{1}Department of Computer Science, ETH Zürich, Switzerland\\
    \textsuperscript{2}Vizrt AG, Zürich, Switzerland\\
    {\tt\small \textsuperscript{1}\{kamanuel,eaksan,jsong,otmarh\}@inf.ethz.ch \textsuperscript{2}\{fpece,rziegler\}@vizrt.com}
    }
    
    \ifappendixonly
            \maketitle
            \thispagestyle{empty}
            \setcounter{figure}{8}

\paragraph*{More motion infilling results.}
\figref{figs:result_zoo} illustrates a number of sequences that we have generated using our method.
Note that all results were produced using the same model, which shows that the model produces plausible and smooth transitions for a variety of different motion types.
Here short sequences selected form the test data where used to force transitions between different motion styles.
Furthermore, we show examples from rare motions that are not well represented in the dataset such as dancing (\ref{figs:result_zoo_c}) and lying on the floor (\ref{figs:result_zoo_d}).
Clearly the model does produce physically infeasible poses such as some floating above the ground before getting up.
However, all results shown are raw predictions without any further processing that would enforce physical plausibility.
\figref{figs:result_zoo_f} shows a sequence composed out of a total of $5$ motion sequences, where two sequences are used to force a grabbing and punching motion respectively.
Please also refer to the video.

\paragraph*{De-noising} \label{sec:denoising}
To demonstrate the de-noising capabilities of the model we perform two sets of experiments (results best seen in the video).
In the first experiment we perturb the joints of the input clip by adding Gaussian noise with zero-mean and unit-variance (see \figref{figs:gaussian_noise_input}). In a second experiment, we mask a joint with a probability of $0.3$ in each frame independently (see \figref{figs:zero_noise_input}).

\begin{figure}[h]
	\begin{subfigure}{0.45\columnwidth}
    	\centering
        \includegraphics[width=\textwidth]{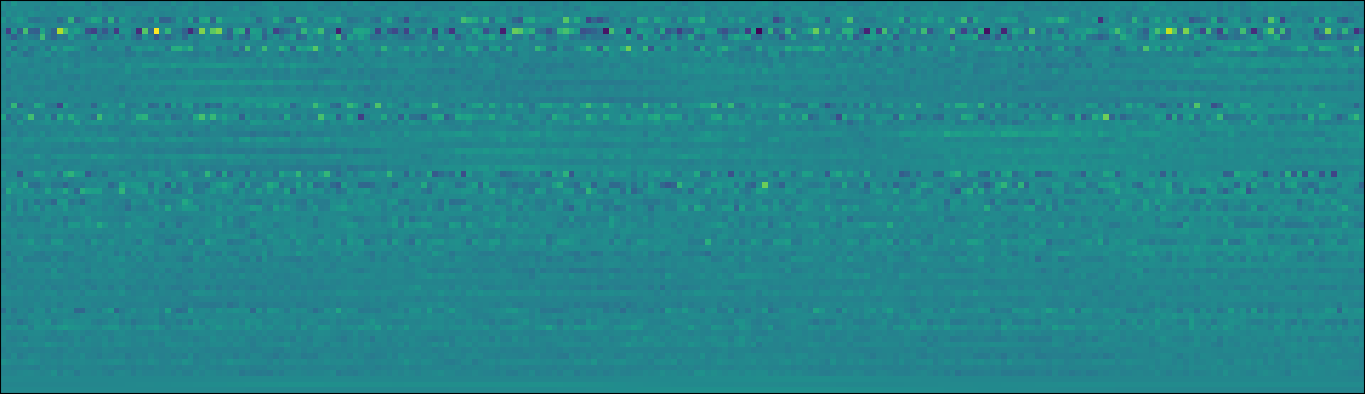}
        \caption{\label{figs:gaussian_noise_input} Gaussian noise.}
    \end{subfigure}
    \hfill
    \begin{subfigure}{.45\columnwidth}
    	\centering
        \includegraphics[width=\textwidth]{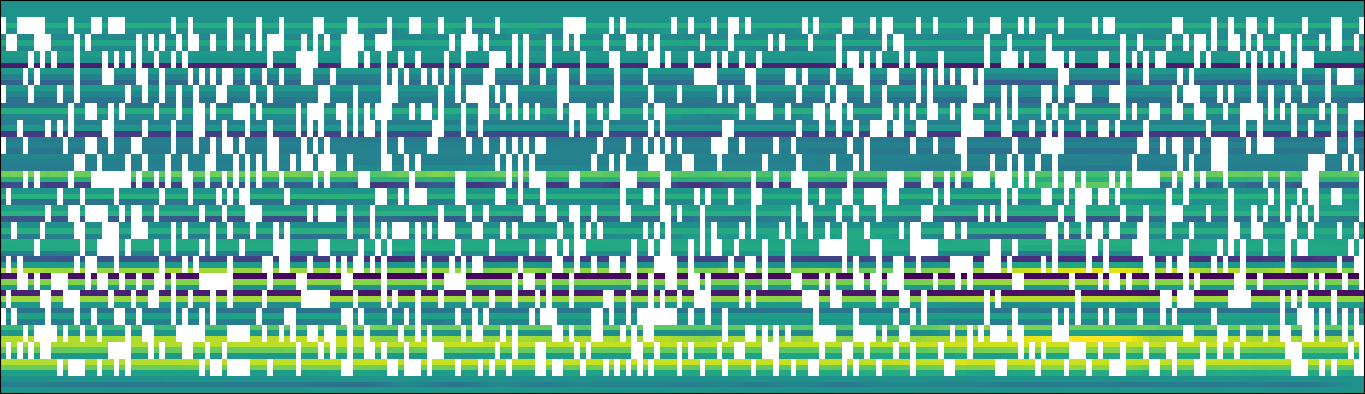}
        \caption{\label{figs:zero_noise_input} Randomly masked joints.}
    \end{subfigure}
	\caption{Sample inputs for our de-noising experiments}
    \label{figs:denoising_input}
\end{figure}

Note that even though our model is not trained specifically for this task, in both experiments it is able to recover a plausible motion out of the corrupted inputs (see \figref{figs:denoise_gauss} and \ref{figs:denoise_zero}). However, we also note that this comes hardly as a surprise since auto-encoder architectures are routinely used for de-noising.

\begin{figure}[h]
    \includegraphics[width=\columnwidth]{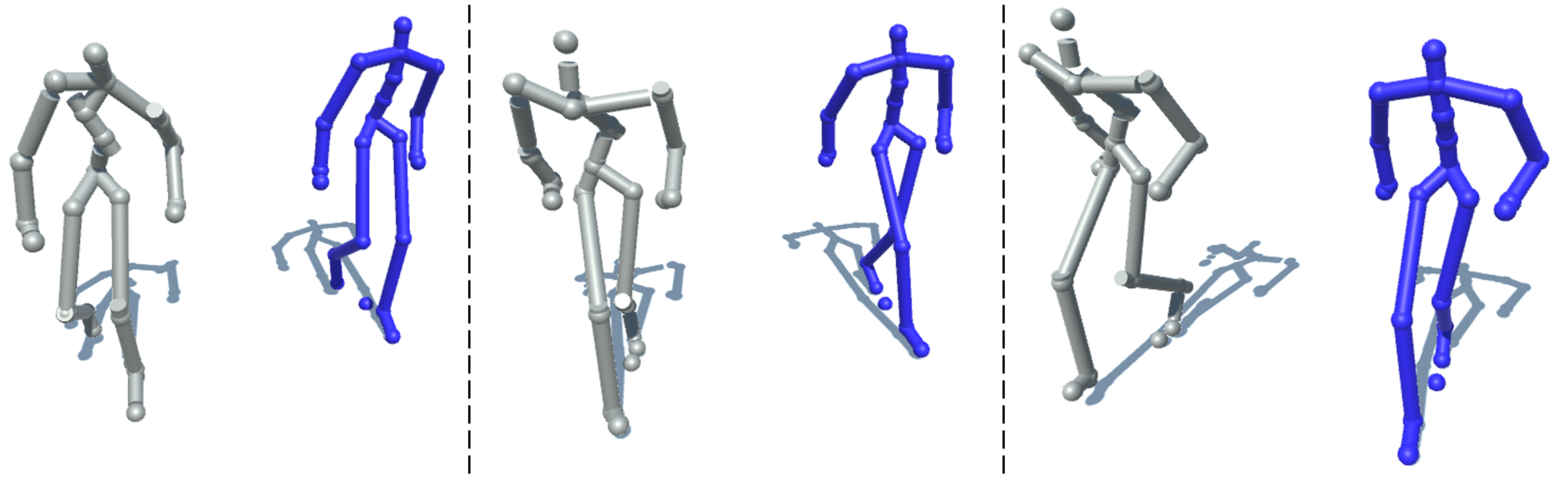}
    \caption{Example of denoising. Gaussian noise on the input data. Perturbed reference in gray and de-noised poses in blue.}
    \label{figs:denoise_gauss}
\end{figure}

\begin{figure}[H]
    \includegraphics[width=\columnwidth]{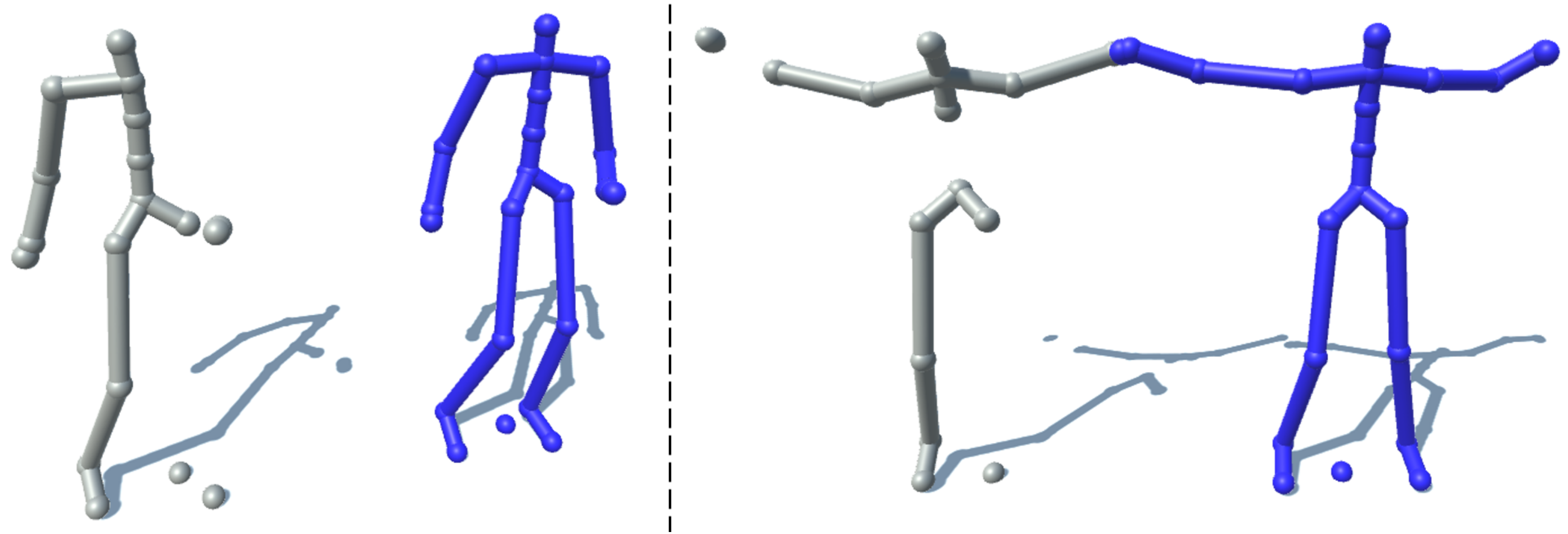}
    \caption{Example of de-noising when in the input a joint is suppressed with probability $p=0.3$ for every frame independently. For example, on the left hand side (in gray), the right wrist and right elbow were removed, while the right hand remained. On the right hand side the entire left leg is missing. In blue is the reconstruction as recovered by our model.}
    \label{figs:denoise_zero}
\end{figure}
\clearpage

\begin{figure*}[h]
	\centering
	\begin{subfigure}{\linewidth}
		\centering
		\includegraphics[width=\linewidth]{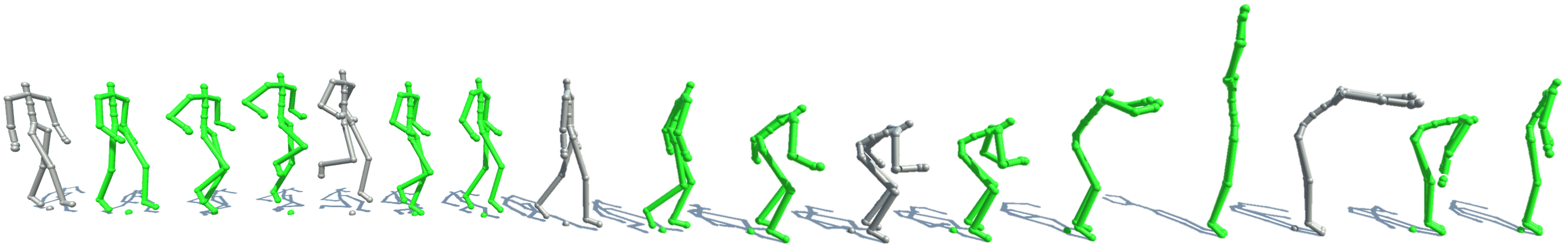}
		\caption{Jumping to bending sequence}
        \label{figs:result_zoo_a}
	\end{subfigure}
	\quad
	\begin{subfigure}{\linewidth}
		\centering
		\includegraphics[width=\linewidth]{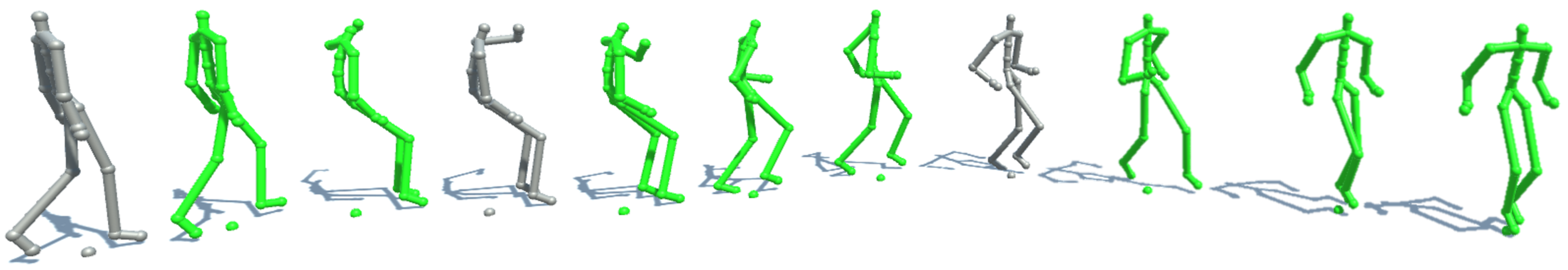}
		\caption{Sitting to running sequence.}
        \label{figs:result_zoo_b}
	\end{subfigure}
    \begin{subfigure}{\linewidth}
		\centering
		\includegraphics[width=\linewidth]{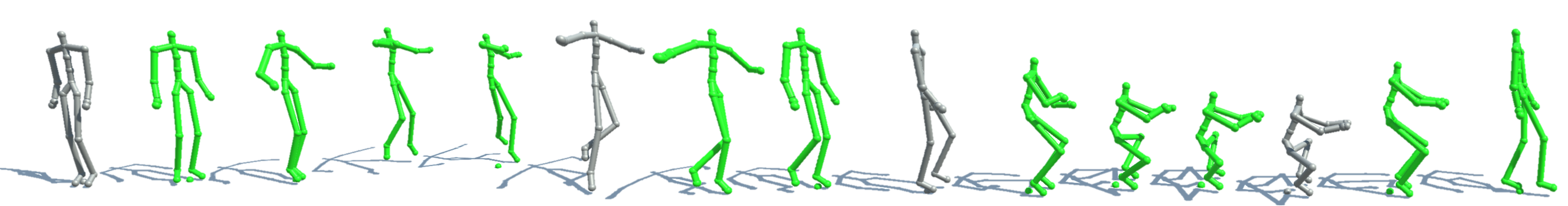}
		\caption{Dancing to squatting sequence.} 
        \label{figs:result_zoo_c}
	\end{subfigure}
    \begin{subfigure}{\linewidth}
		\centering
		\includegraphics[width=\linewidth]{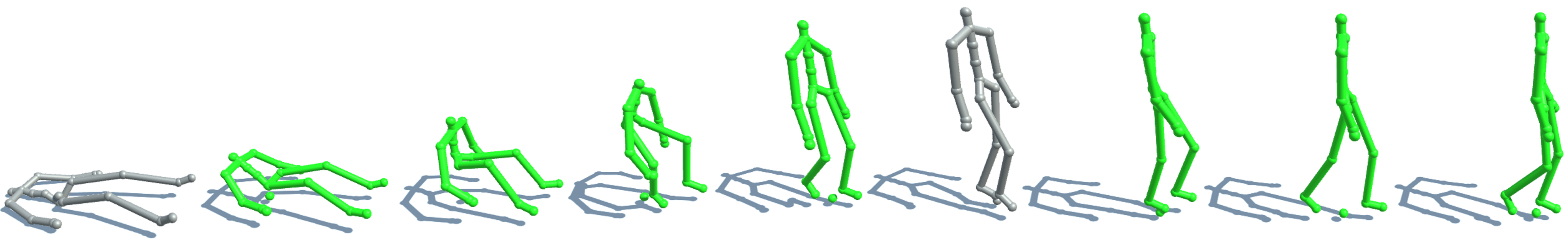}
		\caption{Lying to walking sequence.}
        \label{figs:result_zoo_d}
	\end{subfigure}
    \begin{subfigure}{\linewidth}
		\centering
		\includegraphics[width=\linewidth]{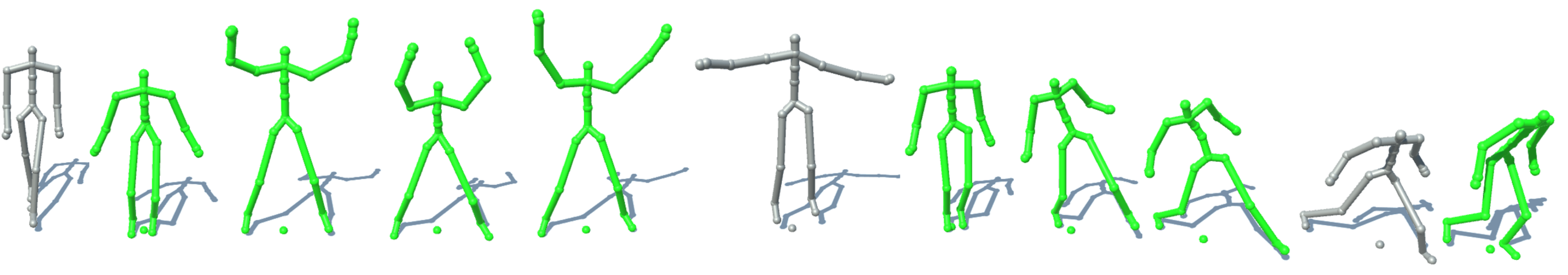}
		\caption{Jumping jacks to large step sequence.}
        \label{figs:result_zoo_e}
	\end{subfigure}
    \begin{subfigure}{\linewidth}
		\centering
		\includegraphics[width=\linewidth]{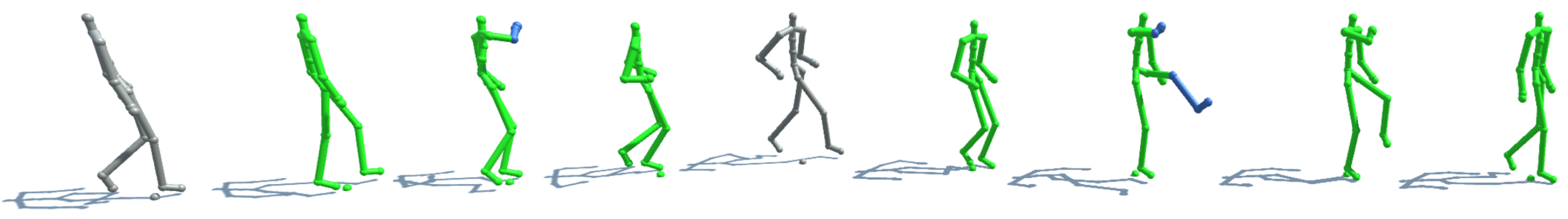}
		\caption{Motion blending of multiple clips. Initially generated sequence in green, additional constraints from different sequences in blue.
		}
        \label{figs:result_zoo_f}
	\end{subfigure}
	    \vspace{0.25cm}
		\caption{More results of motion infilling. Poses in gray correspond to known sequences, while poses in green are filled in by our model.}
    \label{figs:result_zoo}
\end{figure*}

    \else
        \twocolumn[{
                \renewcommand\twocolumn[1][]{#1}%
                \vspace{-3em}
                \maketitle
                \vspace{-3em}
                \begin{center}
            	\centering
            	\includegraphics[width=\textwidth]{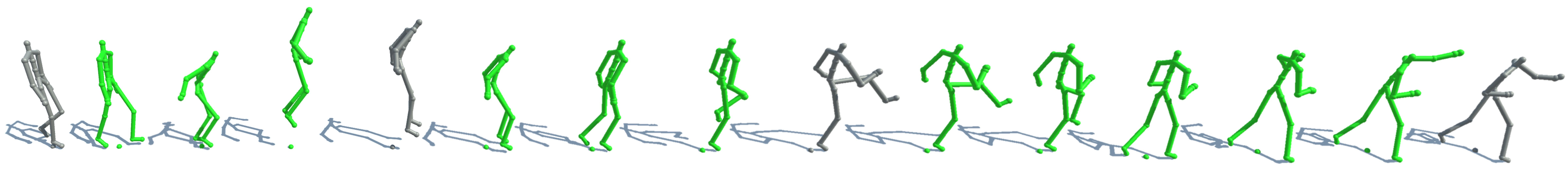}
            	\captionof{figure}{We present a convolutional autoencoder architecture to fill in missing frames in 3D human motion data.
            	Given short sequences of known frames (gray), our method automatically fills in variable-length gaps with realistic and coherent motion data (green).
            	We use a single model to generate motion for a wide range of activities, including locomotion, jumping, kicking, punching, and more.}
                \label{fig:teaser}
                \end{center}
            }]
        \thispagestyle{empty}
        
        \begin{abstract}
            In this paper we propose a convolutional autoencoder to address the problem of motion infilling for 3D human motion data.
            Given a start and end sequence, motion infilling aims to complete the missing gap in between, such that the filled in poses plausibly forecast the start sequence and naturally transition into the end sequence.
            To this end, we propose a single, end-to-end trainable convolutional autoencoder.
            We show that a \emph{single} model can be used to create natural transitions between different types of activities.
            Furthermore, our method is not only able to fill in entire missing frames, but it can also be used to complete gaps where partial poses are available (e.g. from end effectors), or to clean up other forms of noise (e.g. Gaussian).
            Also, the model can fill in an arbitrary number of gaps that potentially vary in length.
            In addition, no further post-processing on the model's outputs is necessary such as smoothing or closing discontinuities at the end of the gap.
            At the heart of our approach lies the idea to cast motion infilling as an inpainting problem and to train a convolutional de-noising autoencoder on image-like representations of motion sequences.
            At training time, blocks of columns are removed from such images and we ask the model to fill in the gaps.
            We demonstrate the versatility of the approach via a number of complex motion sequences and report on thorough evaluations performed to better understand the capabilities and limitations of the proposed approach. 
        \end{abstract}
        

\section{Introduction}
\label{sec:intro}
Modeling 3D human motion has seen increased attention from the computer vision community as it bears the potential to benefit many downstream tasks in robotics, autonomous driving, or human-computer interaction.
Previously, a large body of work has focused on predicting how a given seed sequence evolves in the future \cite{Martinez2017Motion, Ghosh2017, Pavllo2018, Aksan_2019_ICCV, Mao_2019_ICCV, Wang2018Adversarial,Li_2018_CVPR, zhou2018acRNN, Jain:2015:S-RNN}.
While this kind of open-ended motion prediction from past observations has its use cases, there is a significant interest in predicting motion when additionally a sequence in the future is available.
For example, although commercial motion capture systems boast high accuracy, optical tracking systems still struggle with occlusions, especially when multiple people in close contact or objects are involved.
In other cases, due to system failure, damaged tracking sensors, or trackers falling off the actors, valuable data might be lost during a capture session.
Hence, a method to recover missing frames, to complete partially observed poses, or predict missing joints over an entire sequence, is an invaluable tool for post-processing of motion capture data.
Also, keyframe interpolation, a problem that has long been studied in computer graphics, could benefit from such a method.
Automatically filling in large gap sizes is of great interest to animators and artists alike, as it can drastically reduce the amount of required keyframes and hence manual artistic interventions.

This problem of filling in motion between known sequences is sometimes referred to as \textit{motion infilling} and can be defined as completing the gap between an available start sequence in the past and an end sequence in the future (cf. \figref{fig:teaser} or~\ref{figs:ui} for a visual depiction).
Ideally, the gap in between is completed in such a way that the resulting motion is plausible, natural, and the transitions at either end of the gap are smooth.
Motion infilling can also refer to the case where a few joints are missing over the entirety of a sequence or when only partial poses are observed.

In this paper we present a simple yet effective convolutional autoencoder (CAE) trained to complete long gaps in human motion sequences producing compelling and smooth transitions without further post-processing steps. Our proposed model is fully convolutional and able to synthesize motion for varying gap sizes and multiple gaps occurring in the same sequence. Furthermore, it can also be used to blend together partially observed poses within the gap, or single joints missing over the entire sequence. The proposed method lends itself naturally to the removal of other types of noise as well, for example Gaussian noise, and as such can be viewed as a general de-noising framework.

Motion infilling is challenging as it requires learning smooth transitions between possibly different types of motion.
Also, recent success in open-ended motion prediction does not immediately translate to the infilling task, as the dominant method of choice has been recurrent neural networks (RNNs), which are notoriously difficult to condition to future sequences.
For example, Berglund \etal \cite{Berglund2015Birnn} show how a bi-directional structure can be used for the task, but the supported gap size is fixed and short.  
Very recently, Harvey \etal \cite{Harvey:2020:robustInbetweeningL} introduced time-to-arrival embeddings to a recurrent structure to support variable gap sizes.
However, the authors also report the use of inverse kinematics to post-process the model's outputs.

To avoid the limitations of recurrent models in motion modeling, we suggest to cast motion infilling as an image inpainting problem.
Here, the ``image'' corresponds to a matrix-based representation of the motion data, where each column contains a single pose and each row stands for a joint's information over time. Treating motion sequences as matrices has been studied before, e.g. \cite{Butepage2017RepL, Holden:2015:LMM,Hernandez_2019_ICCV, Li_2018_CVPR}. 
The problem arising when using such representations is that neighboring ``pixels'' are not necessarily neighboring joints in the skeletal hierarchy.
Hence, capturing the intricate spatial dependencies is not straightforward, especially for convolutional architectures that usually excel at exploiting spatial relationships from images.
Previous work addresses this by using dense layers spanning the entire height of the input matrix \cite{Butepage2017RepL, Hernandez_2019_ICCV}, using large kernel sizes \cite{Li_2018_CVPR}, or resorting to 1D convolutions over the temporal domain only \cite{Holden:2015:LMM}.

In this work we offer a different take on this problem and propose that a model with a sufficiently large receptive field is able to capture the spatial relationships.
We achieve this by designing a convolutional auto-encoder architecture inspired by the VGG model \cite{Simonyan:2014:VDP}.
This way, the model covers a large number of joints in the upper layers and is furthermore capable of learning temporal dependencies to produce smooth motions without collapsing to a safe mean pose.
We train our model by using a reconstruction objective only and do not require any additional regularization terms or adversarial training. To enable variable-length infilling, we employ a curriculum learning scheme in which the inputs are perturbed with increasingly large gaps during training.
This results in a model that can easily fill in up to 120 frames (2 seconds) between short given sequences (i.e., typically 30--40 frames but also as little as 5). 

Our proposed model allows us to train a \textit{single} model encompassing a large variety of motion types like walking, jumping, kicking and more, while still fulfilling the constraints defined by the known sequences as closely as possible and without requiring post-processing. In doing so we show that a simple but effective convolutional model can indeed be used to unravel the complex spatio-temporal dependencies in 3D human motion data.

In summary, this paper contributes an easy-to-train and efficient CAE for the task of 3D human motion infilling.
The proposed model is able to synthesize convincing looking motion to fill in variable-length gaps between sparsely distributed known sequences of various motion types.
Unlike previous work we push the boundaries of generating smooth and natural looking transitions using the network's outputs directly and thus foregoing post-processing steps or regularization terms.
The resulting model can be applied to a variety of infilling and other noise removal tasks as demonstrated by our evaluations.


\begin{figure}[t]
\includegraphics[width=\columnwidth, trim=0 0 1100 0, clip]{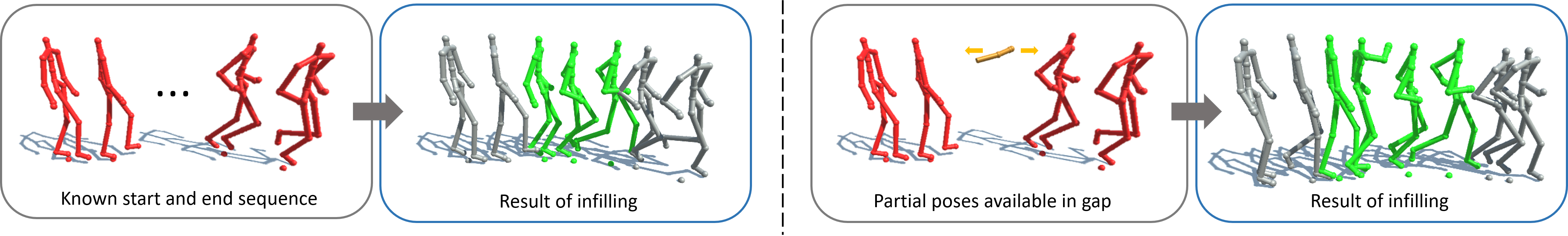}
\caption{Given a known start and target sequence with a gap inbetween, our model fills in the missing frames (shown in green) creating a smooth transition from walking to running. Note that while only a single gap is shown here, the proposed method is able to complete several gaps of varying size.}
\label{figs:ui}
\end{figure}

\section{Related Work}
\label{sec:rw}

\paragraph*{Recurrent Models}
Modeling human motion has been of interest to the machine learning community as an instance of time-series modeling, with early works by Taylor \etal \cite{Taylor2006, Taylor:2011:TDM} employing conditional restricted Boltzmann machines.
In more recent years, the computer vision community has often adopted RNNs to address the task of predicting motion into the future given a seed sequence \cite{Fragkiadaki:2015:RNM, Jain:2015:S-RNN, Ghosh2017, Aksan_2019_ICCV, Pavllo2018, Martinez2017Motion, zhou2018acRNN, Wang2018Adversarial}.
To avoid a collapse to a mean pose, researchers have suggested to perturb the input at test time with Gaussian noise \cite{Fragkiadaki:2015:RNM}, removing joints entirely \cite{Ghosh2017}, training the model on its own outputs \cite{Martinez2017Motion}, or employing a specialized output layer that follows the kinematic chain of the skeleton \cite{Aksan_2019_ICCV}.
Martinez \etal \cite{Martinez2017Motion} propose a residual connection in a sequence-to-sequence architecture to address the discontinuity between the last known and first predicted pose. Wang \etal \cite{Wang2018Adversarial} further improve upon smoothness and realism of the predictions by employing adversarial and geodesic-inspired losses.

While recurrent models lend themselves well to open-ended motion prediction, they are less suitable for the task of motion infilling studied in this paper.
An example of a recurrent structure used for infilling of joint trajectories is presented by Berglund \etal \cite{Berglund:2015:BIRNN} in the form of a bidirectional RNN.
However, the presented method uses a fixed gap size, is only demonstrated for small gaps, and is rather expensive.
Only recently Harvey \etal \cite{Harvey2019RTN} have shown the application of a recurrent structure to motion infilling for larger gap sizes.
However, in \cite{Harvey2019RTN} the authors mention that post-optimization is used to close discontinuities in the transition from the last predicted frame to the first known frame of the target sequence.
\cite{Harvey:2020:robustInbetweeningL} is a recent follow-up work. While demonstrating impressive results, it mentions the use of inverse kinematics as a post-processing step.
In this work, we focus on exploring the capabilities of a neural network alone to produce smooth and natural motion for the task of infilling, i.e. without relying on regularization or post-processing.

\paragraph*{Non-recurrent Models}
While non-recurrent models have prominently been used for time-series modeling of audio or natural language (e.g., \cite{Dauphin2018, oord2016wavenet,aksan2019stcn}), their adoption for human motion data is not straightforward.
This can be explained by the fact that in addition to temporal dependencies, also highly complex spatial dependencies related to the dynamics of human motion must be captured.

Mao \etal \cite{Mao_2019_ICCV} were the first to apply a graph convolutional network (GCN) to the task of motion prediction.
The GCN operates on discrete cosine transforms of the input data, which makes it more difficult to support variable length inputs.
In \cite{Butepage2017RepL}, B\"utepage \etal also represent the data as matrices similar to our work. They then use dense layers in an autoencoder framework to learn meaningful representations, which can be leveraged for prediction and infilling. In contrast to our work, their dense layers span the entire spatial domain of the input matrix, whereas we use small convolutional filters. This allows us to harness the efficiency and computational advantage of CNNs, while still capturing the spatio-temporal dependencies of the data.

Convolutional models have been applied to motion prediction by Hernandez \etal \cite{Hernandez_2019_ICCV} and Li \etal \cite{Li_2018_CVPR}.
Both works represent motion sequences in matrices, like we do in this work.
To apply convolutions to this data representation, all approaches must deal with the fact that neighboring values in the data matrix are not necessarily neighboring joints in the human skeleton \cite{Hernandez_2019_ICCV}.
\cite{Hernandez_2019_ICCV} do so by employing dense layers that span the entire height of the input matrix.
While we agree with \cite{Hernandez_2019_ICCV} that the lack of spatial proximity in the inputs is an additional challenge for any CNN, we show in this work that with a sufficiently big receptive field, this is no hindrance for successful motion modeling with CNNs.
Given a deep enough model, a CNN with 2D filters spanning both spatial and temporal dimensions is indeed able to unfold the underlying spatio-temporal relationships. 

Like in this work, Li \etal \cite{Li_2018_CVPR} use convolutions directly on the input matrix. However, they use bigger filter sizes and do not apply their architecture to motion infilling.
Also, we refrain from using any adversarial losses or other kinds of regularizers (e.g. bone length constraints in \cite{Hernandez_2019_ICCV}) and instead demonstrate that our model discovers desirable properties such as bone length consistency and smoothness from the data alone without external guidance other than the reconstruction loss.

\paragraph*{Manifold Learning and De-noising}
CNNs have been used to learn the projection of human motion onto a low-dimensional manifold.
Drawing samples from this manifold then generates plausible human poses.
Holden \etal \cite{Holden:2015:LMM} use such an approach to recover full poses from partial inputs such as corrupted motion-capture data and to fill gaps of fixed sizes (i.e., max 15 frames).
A custom un-pooling operation requires careful layer-by-layer training.
In follow-up work \cite{Holden:2016:DLF} a similar but shallower architecture is integrated into a hierarchical system that maps from foot-fall patterns and trajectories to pose sequences.
However, a separate model has to be trained for each activity, thus this approach is not well suited to interpolate between sequences of different activities.
We also leverage convolutional auto-encoders but train a single model for all motion types, predict both poses and rotational and translational velocities and can handle transitions between different activities.

As our method cannot only be used to fill in missing frames, but also to clean up other types of noise such as Gaussian, it is related to the work of Holden \cite{Holden2018Denoising}.
Although Holden's architecture operates on marker level and not joints, our method is in theory directly applicable.
In contrast to our work, \cite{Holden2018Denoising} uses a frame-wise ResNet-based architecture and requires post-optimization to induce smoothness and improved pose reconstruction.

\paragraph*{Motion Modeling in Computer Graphics}
A variant of motion infilling, known as keyframing, is a long-standing problem in the computer graphics community with the latest addition provided by \cite{Harvey:2020:robustInbetweeningL}.
Keyframing aims at reducing the amount of tedious manual labor for artists to create life-like animations without relinquishing control over the resulting animation.
Early works date back to the 1980s \cite{Witkin:1988:SC} and many other approaches with the same goal have been proposed since. Such approaches fall into the field of physics-based character animation \cite{van1997footprints, Agrawal:2016:TBL}, space-time optimization \cite{abe2006momentum,liu2002synthesis,Mordatch:2012:DCB} or motion graphs \cite{Kovar:2002:MG,Yamane:2010:PSS, Min:2012:MGC}.
While our method can be understood as a keyframing or general character animation tool, in this work we do not intend to provide a production-ready tool.
Instead, we focus on the more basic building block of synthesizing convincing human motion from neural networks alone.

\paragraph*{Image Inpainting}
Our approach is inspired by work on image inpainting \cite{Yang:2016:HRII, Pathak:2016:FLI, Yeh:2016:SII}, proposing variations of the idea to combine $\ell_1$ or $\ell_2$ penalized auto-encoders with an adversarial loss in order to synthesize realistic looking natural images.
We cast motion infilling as an image inpainting task and also leverage an auto-encoder approach. However, the idiosyncrasies of the task and the differences in the data require a custom designed architecture.


\section{Method}
\label{sec:method}

\begin{figure}
\includegraphics[width=\columnwidth]{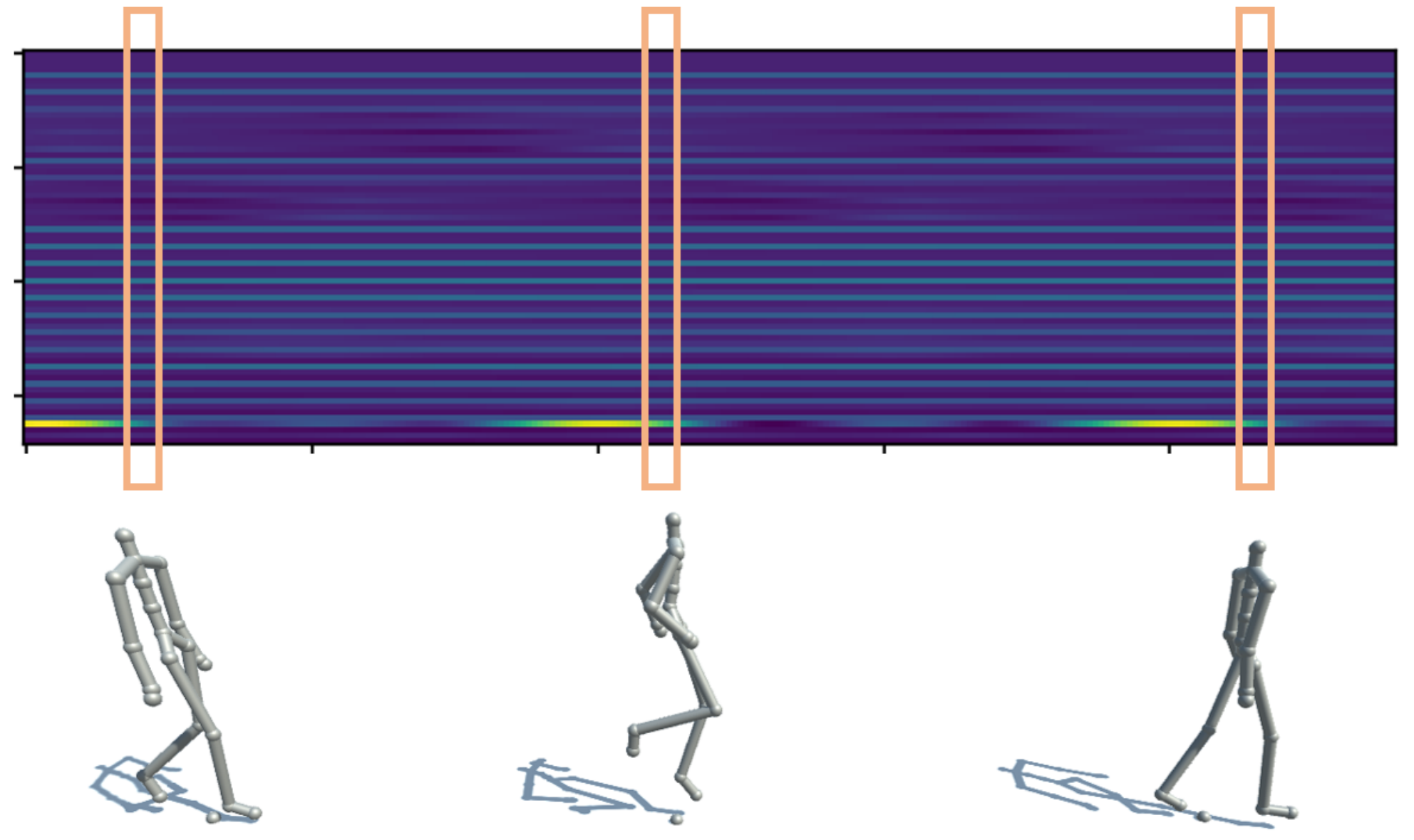}
\caption{The input to our model consists of motion clips represented as matrices that can be viewed as images.
Each column of the image corresponds to a particular frame in the sequence (highlighted in orange), while rows encode joint positions over time. The root-relative 3D joint positions are flattened into a single vector. To each pose vector we also append translational and rotational velocity information of the root.}
\label{figs:data_representation}
\end{figure}

In this section we detail the architecture, training, and inference procedure for our model.
We propose a deep convolutional de-noising auto-encoder $\mathcal{M}$, trained to fill in missing 3D pose information represented as image-like joint position maps (cf. \figref{figs:data_representation}).
That is, given the constraints for joint positions at the beginning and end of a motion sequence, our model reconstructs the missing frames in between in a visually coherent way.

We cast the infilling problem as an image inpainting task, where the image corresponds to the joint position map.
To do so, we must address the problem that joints that are close in image space are not necessarily spatially close in the skeleton hierarchy, which seems to go against the basic assumption of CNNs.
Nonetheless, we show that given a sufficiently deep model and sufficiently large receptive field, a convolutional autoencoder with small filter sizes in the spatial and temporal domain is indeed able to capture the underlying spatio-temporal structure of the data.
We achieve this by following well-established design rules, i.e. stacking several convolutional layers (with $3 \times 3$ filters) and pooling operations inbetween.
This is in contrast to previous work that either resorts to dense layers to encode the spatial relationships \cite{Butepage2017RepL, Hernandez_2019_ICCV}, uses larger filter sizes \cite{Li_2018_CVPR} or temporal convolutions only \cite{Holden:2015:LMM}.

Furthermore, in order to forgo an explicit modelling of the skeletal structure we propose a  curriculum learning training scheme that aids the model in learning the spatial dependencies between joints, via randomized removal of joint information.
In consequence, the model is robust against different types of perturbations in the inputs and thus it can be used for different tasks besides motion infilling, such as noise reduction.
We show various applications of our method in \secref{sec:eval} and in the accompanying video.

\subsection{Data Representation}\label{subsec:data-representation}
Here, we briefly introduce the data representation used as input to our model. For an overview please refer to \figref{figs:data_representation}.

We utilize the dataset introduced by Holden \etal \cite{Holden:2016:DLF}, which in turn is a collection of publicly available datasets \cite{CMU,MHAD,HDM05,Holden:2015:LMM}.
This dataset represents poses as 3D positions of joints in space.
From this dataset we extract motion clips $\boldsymbol{X}_i \in \mathbb{R}^{P \times T}$, where $T$ is the number of frames and $P$ the size of the vector representing the pose at time step $t$.
For training, we fix $T$ to $240$ ($4$ seconds), but at inference time it is variable due to the fully convolutional nature of our model.
This results in a dataset that contains $26'088$ clips extracted from $5'648$ unique sequences, of which we reserve $90 \%$ for training.

Like \cite{Holden:2016:DLF}, we use a local body coordinate format to represent a pose at time step $t$.
To this end, the $(x, y, z)$ positions of each joint at frame $t$ are flattened and concatenated to form a pose vector.
The joint positions are all relative to the position of the root.
In addition, translational velocities $\boldsymbol{t} \in \mathbb{R}^2$ in the floor plane are extracted from the root trajectory, as well as a rotational velocity $\gamma \in \mathbb{R}$ about the up-axis.
This representation of the root trajectory $\boldsymbol{r} = \left[ \boldsymbol{t}, \gamma \right]$ is always relative to the previous frame and we concatenate it to the flattened vector of joint positions.
Using 22 joints, this means that the pose at step $t$ can be represented as a $69$-dimensional vector, i.e. $P = 69$.

\subsection{Model Architecture}\label{sec:model}

\begin{figure}[t]

\begin{subfigure}{.48\textwidth}
    \includegraphics[width=\columnwidth]{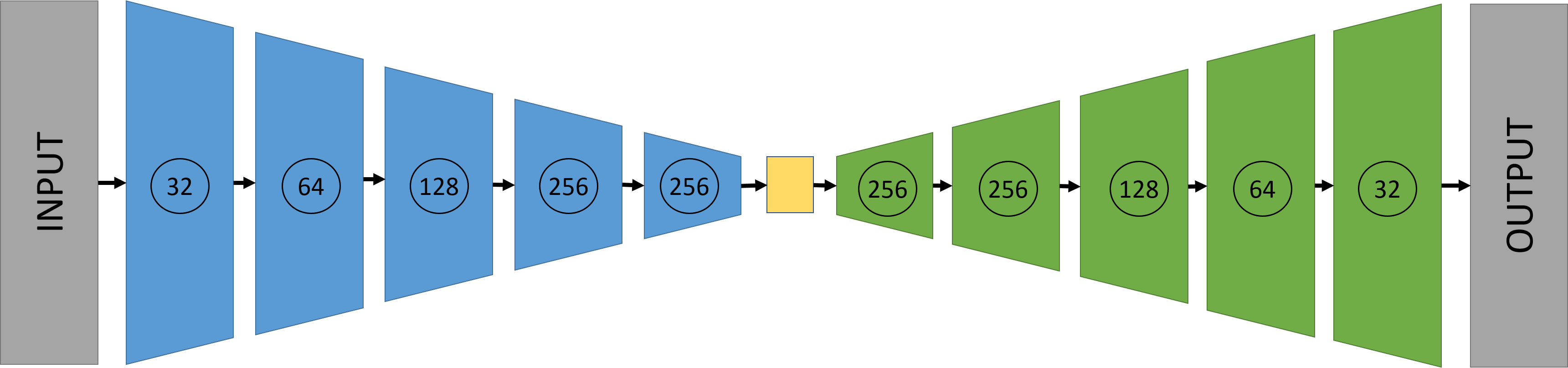}
    \caption{Our model contains 5 layers of encoding and decoding units. The circled numbers indicate the number of output channels.}
    \label{figs:vgg_aec_network}
\end{subfigure}
\begin{subfigure}{.48\textwidth}
    \includegraphics[width=\columnwidth]{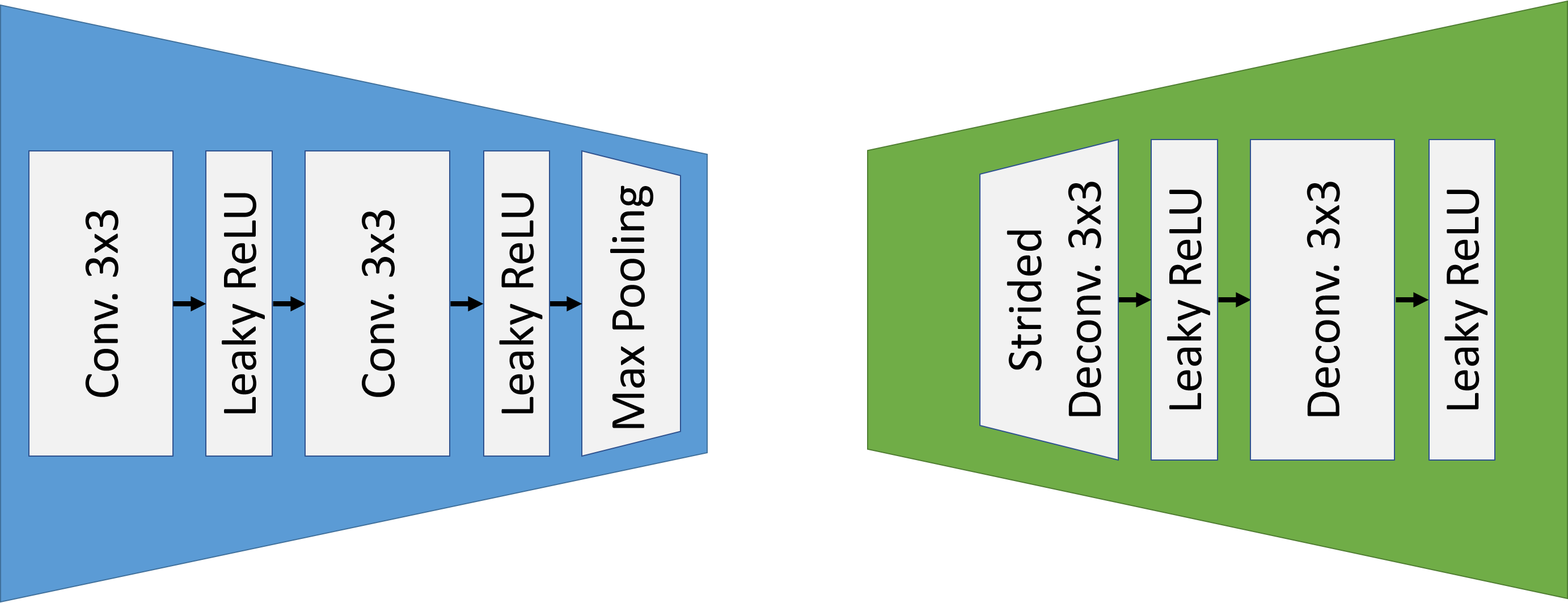}
    \caption{Encoding (left) and decoding (right) unit. There is no activation on the output layer of the last decoding unit.}
    \label{figs:vgg_units}
\end{subfigure}
\vspace{0.1cm}
\caption{Architecture overview of our model.}
\label{fig:vgg_ae}
\end{figure}

Our proposed architecture is inspired by \cite{Holden:2016:DLF} and the VGG model design \cite{Simonyan:2014:VDP}.
An overview is illustrated in \figref{fig:vgg_ae}.
The main reasons speaking for such a deep architecture are larger model capacity, more non-linearities to model inherent complexities of the task and increasingly large receptive fields.
The latter allows the model to implicitly learn dependencies across large spatial distances (in image space) \cite{Simonyan:2014:VDP}.
In our case this means that the model can implicitly recover the joint dependencies in the skeleton without requiring a pre-determined spatial prior.

As shown in \figref{figs:vgg_aec_network}, the encoder network stacks 5 encoding units, where each unit consists of two $3 \times 3$ convolutional layers whose outputs are activated using a leaky ReLU.
Like in the VGG model, the encoding units (cf. \figref{figs:vgg_units}, left) are designed to increase the receptive field while keeping the increase in trainable parameters minimal.
Via max-pooling, the output of each encoding unit is halved in both the spatial and temporal dimension before being passed into the subsequent unit.
The decoder network mirrors the architecture of the encoder (\figref{figs:vgg_units}, right), but the two are decoupled, i.e. do not share weights.
To revert the pooling operation in the decoder, we use strided de-convolutions.

While our model looks similar to the motion manifold network proposed by Holden \etal, we would like to highlight the following two key differences.
First, the autoencoder layers proposed by \cite{Holden:2015:LMM, Holden:2016:DLF} use 1D convolutions over the temporal domain, while we employ small 2D filters.
Second, \cite{Holden:2015:LMM, Holden:2016:DLF} propose a custom un-pooling operation, denoted as $\Psi^\dagger$ in \cite{Holden:2016:DLF}.
In our experiments, stacking several of those layers has proven to be difficult to train in an end-to-end fashion.
In \cite{Holden:2015:LMM} layer-wise training is employed to train the autoencoder.
Still, \cite{Holden:2016:DLF} reports that, even if trained successfully, stacking multiple layers blurs out the results and hence the authors fall back to a single-layer autoencoder.
Our model uses strided de-convolutions to replace $\Psi^\dagger$, which enables straight-forward end-to-end training of multi-layer architectures and leads to improved performance (cf. \secref{sec:eval_quant}).

For a typical input sample of size $69 \times 240$, the resulting embedding in the latent space (yellow box in \figref{figs:vgg_aec_network}) corresponds to a $3 \times 8 \times 256$ tensor.
Obviously the question whether such a compact representation can accurately represent the subspace of valid poses arises.
To this end we refer to our experimental results in \secref{sec:eval} indicating that our architecture can indeed model the sub-space well.
Moreover, due to the deep architecture the effective size in the upper layers is comparatively large and hence contributes to capturing long-range relationships.

\subsection{Training}
\label{sec:training}
The task of motion infilling is now cast as the problem to fill in missing gaps within the ``images'' $\boldsymbol{X}_i$.
Note that this is different from the usual application scenario of de-noising autoencoders, where inputs are typically assumed to be corrupted by some stochastic diffusion process.
In our case we remove entire blocks of complete columns from the inputs (see \figref{figs:inbetweening_input}) so that the model is forced to learn how to generate complete, valid poses.

During training we present the model with pairs of complete and incomplete samples, and we instruct it to reconstruct the data to closely match the complete counterpart.
To induce robustness against the amount of perturbation we employ a curriculum learning training scheme \cite{Bengio:2009:CL}.
During curriculum learning the model is initially trained with samples that allow the task to be easily completed, and subsequently it is exposed to increasingly harder cases.
In our case, this translates into increasing the span of masked ranges along the temporal dimension.
In this way the model is exposed to more realistic representations of the problem, and consequently is capable of generating plausible poses given drastically different types of inputs.
In particular, the model is robust to the length of the gap, to random noise applied to inputs and to missing joint information.

More formally, during training we generate corrupted motion samples by perturbing the clips $\boldsymbol{X}_i$ with the help of a mask $\boldsymbol{M}_i \in \mathbb{R}^{P \times T}$ as follows:
\begin{equation*}
\begin{split}
\tilde{\boldsymbol{X}}_i = \boldsymbol{X}_i \odot \boldsymbol{M}_i
\end{split}
\end{equation*}
where $\odot$ denotes the element-wise matrix multiplication.
The model $\mathcal{M}$ then tries to reconstruct the perturbed motion samples $\tilde{\boldsymbol{X}}_i$ such that:
\begin{equation*}
\begin{split}
\hat{\boldsymbol{X}}_i = \mathcal{M}(\tilde{\boldsymbol{X}}_i).
\end{split}
\end{equation*}

We parameterize the choice of the mask $\boldsymbol{M}_i$ by a pair of scalars $(\lambda, \tau)$, where $\lambda$ and $\tau$ determine the length and location of the gap, respectively.
In order to increase the variation of the inputs we generate the masks by using normal and uniform distributions such that:
\begin{equation*}
\lambda \sim \mathcal{N}({\mu}_e, {\sigma}^2) \quad \quad
\tau \sim \mathcal{U}(1, T-\lambda)
\end{equation*}
where ${\mu}_e$ and ${\sigma}^2$ parametrize the gap size distribution at training epoch $e$. Please note, that while ${\mu}_e$ is gradually increased during training, ${\sigma}^2$ is kept fixed. We initialize $\mu_e$ to $10$, and increase it until $120$ by steps of $10$ every $5$ epochs.
Additionally, we randomly alternate between the previously described type of distortion and masking out $1, 2,$ or $3$ random joints over the entire length of the sequence.

For each pair of original and reconstructed samples $(\boldsymbol{X}_i, \hat{\boldsymbol{X}}_i)$, we optimize the $\ell_1$ loss: 
\begin{equation*}
\mathcal{L}(\cdot) = \norm {\boldsymbol{X}_i - \hat{\boldsymbol{X}}_i}_1
\end{equation*}
In our experiments we found that the $\ell_1$ loss results in smoother reconstructions when compared to the $\ell_2$ loss, which reflects previous work \cite{Zhao:2015:LOSS} on image inpainting.
Please note that we compute the loss over the entire motion sequence, not just the masked portion.
In this way we force the model to learn the global structure of the motion.

We apply zero-mean and unit-variance normalization on each row of $\boldsymbol{X}_i$, where the mean and standard deviation are calculated over the training dataset.
We use the Adam optimizer \cite{Kingma:14:ADAM} with a fixed learning rate of $0.001$ and a batch size of $80$.
For all convolutional layers, Xavier initialization is used \cite{Glorot:2010:XINIT}.
We train for $200$ epochs which roughly takes 8 hours on an NVIDIA Quadro M6000 (12 GB).


\section{Evaluation}
\label{sec:eval}
In this section, we show quantitative comparisons and ablations (\secref{sec:eval_quant}) and demonstrate the visual quality of our method (\secref{sec:eval_qual}).
Please also refer to the video for more visualizations.

\subsection{Quantitative Results}
\label{sec:eval_quant}

\paragraph*{Reconstruction and Infilling}
To assess our method's performance quantitatively we compute the mean 3D joint error for both reconstruction and infilling tasks.
\tabref{tab:infilling_results} compares our best model to three baselines:
\textit{Interpolation} refers to naive linear interpolation.
\textit{Vanilla AE} is the same as our best model but each encoding/decoding block contains only a single strided $3 \times 3$ convolutional layer, yielding a smaller receptive field.
We also compare against an implementation of the motion manifold network presented in \cite{Holden:2016:DLF}.

Finally, we also show the effect of curriculum learning on our best model. While the previous baselines were trained to reconstruct 60 frames (1 second) in clips of size 240, the last entry in \tabref{tab:infilling_results} was trained to reconstruct up to 120 frames (2 seconds) with the curriculum scheme described in \secref{sec:training}.

\tabref{tab:infilling_results} shows that a simple application of \cite{Holden:2016:DLF} is suboptimal.
Our architectural changes described in \secref{sec:model} lead to improved performance (\textit{Vanilla AE, Ours}).
Furthermore, the difference in performance between the \textit{Vanilla AE} and our model highlights the importance of the receptive field's size.
Lastly, the model trained using the curriculum loses some accuracy in the 60-frame and 0-frame (i.e., pure reconstruction) tasks. Despite this trade-off, \textit{Ours (curr.)} still ranks second at least.

\begin{table}[t]
    \centering
    \resizebox{\linewidth}{!}{%
    \begin{tabular}{l|c|c|c|}
         Model & 0 Frames & 60 Frames & 120 Frames \\
         \hline
         Linear Interpolation & n/a & 10.45 ($\pm$ 15.5) & 17.04 ($\pm$ 24.4)\\
         Holden \etal (60) \cite{Holden:2016:DLF}  & 4.44 ($\pm$ 6.5) & 15.28 ($\pm$ 19.1) & 18.26 ($\pm$ 24.5)\\
         Vanilla AE (60) & 2.79 ($\pm$ 6.7) & 7.45 ($\pm$ 11.7) & 13.61 ($\pm$ 21.6)\\
         Ours (60) & \textbf{2.37} ($\pm$ 5.0) & \textbf{4.96} ($\pm$ 8.5) & 12.00 ($\pm$ 19.5) \\
         Ours (curr.) & 2.75 ($\pm$ 5.2) & 5.47 ($\pm$ 8.6) & \textbf{7.00} ($\pm$ 10.6) \\
    \end{tabular}}
    \vspace{0.1 cm}
    \caption{3D joint error comparison for reconstruction (\textit{0 Frames}) and infilling (\textit{60/120 frames}) tasks. Reported are mean and standard deviation in cm computed over all joints and frames in the validation set. Poses are root aligned. For infilling results the error is computed only over the gap. \textit{(60)} after the model name means that this model was trained to reconstruct 60 frames. \textit{(curr.)} refers to curriculum learning up to 120 frames.}
    \label{tab:infilling_results}
\end{table}

\begin{table}[t]
    \centering
    \resizebox{\linewidth}{!}{%
    \begin{tabular}{l|c|c||c|c|}
         Model & $\sigma = 1.0$ & $\sigma = 1.5$ & $p = 0.3$ & $p = 0.5$ \\
         \hline
         Holden \etal \cite{Holden:2016:DLF} & \textbf{5.0} ($\pm$ 4.8) & \textbf{6.6} ($\pm$ 5.4) & 6.5 ($\pm$ 7.8) & 9.5 ($\pm$ 11.3)\\
         Ours & 5.8 ($\pm$ 4.6) & 7.8 ($\pm$ 5.9) & \textbf{3.1} ($\pm$ 4.4) & \textbf{4.0} ($\pm$ 5.6)\\
    \end{tabular}}
    \vspace{0.1 cm}
    \caption{3D joint error comparison for de-noising experiments. $\sigma$ indicates how much Gaussian noise was added to the inputs. $p$ is the percentage of joints dropped at random in each frame. Values are mean and standard deviation in cm.}
    \label{tab:denoising_results}
\end{table}

\paragraph*{De-noising}
To test our method's noise reduction capabilities, we conduct two experiments.
In the first one we perturb the inputs by adding Gaussian noise $\epsilon \sim \mathcal{N}(0, \sigma)$ to the joint positions.
In a second experiment, we mask a joint with a probability of $p$ in each frame independently.
\tabref{tab:denoising_results} summarizes the 3D joint reconstruction error for varying values of $\sigma$ and $p$ computed over 400 randomly chosen validation samples of length 240 frames.
In removing Gaussian noise from the inputs, our best model is out-performed by the baseline. However, our model beats the baseline by a large margin when filling in randomly dropped joints.
Note that neither models were trained specifically for these tasks.
Furthermore, the 3D joint reconstruction metric does not necessarily correlate well with perceived smoothness or naturalness of the output motion.
In the supplementary video we visualize several de-noised sequences and show that our model's outputs are both smooth and plausible.

\paragraph*{Bone Length Consistency}
Our method does not use an auxiliary loss to enforce bone length consistency as we have found this to have little effect on the overall results.
To investigate how much bone length variation we observe in the model outputs, we compute the bone lengths over the entire validation set for the pure reconstruction task.
\figref{figs:bone_length} compares our best model to our implementation of Holden \etal \cite{Holden:2016:DLF}.
While our model clearly exhibits variance, the spread of predicted bone lengths is smaller compared to \cite{Holden:2016:DLF} and the median of all predictions is typically closer to the ground-truth value.

\begin{figure}
\centering
\includegraphics[width=0.48\textwidth]{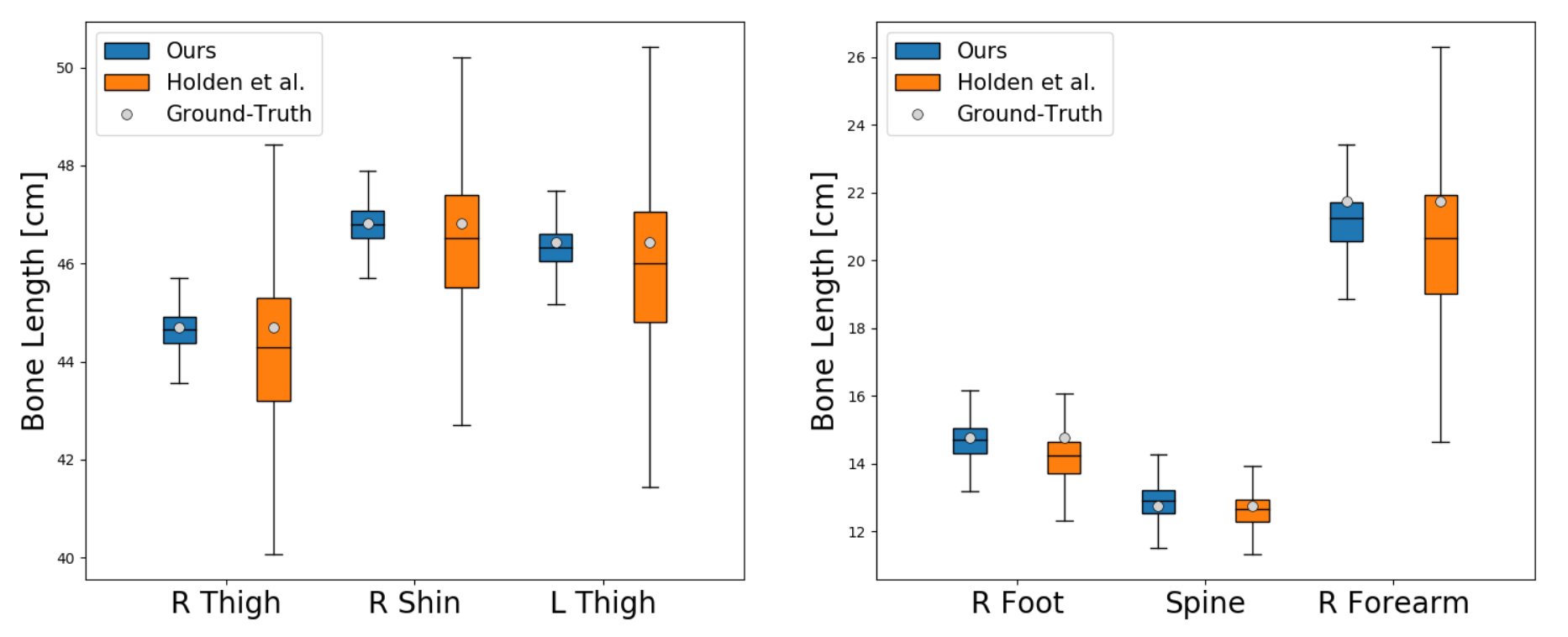}
\caption{\label{figs:bone_length} Boxplots of estimated bone lengths for 6 selected bones computed over the entire validation set ($n = 610'080$) when performing pure reconstruction. We compare our best model (in blue) vs Holden \etal \cite{Holden:2016:DLF} (in orange) vs ground-truth values (gray dots).}
\end{figure}

\subsection{Qualitative Results}
\label{sec:eval_qual}
In the following we present our experiments assessing the quality of the synthesized motions.
Please note that all figures and the accompanying video show raw model outputs without any additional post-processing.
Furthermore, the model predicts the entire sequence, i.e. the known subsequences are replaced with reconstructed poses.
This allows the model to slightly alter the given frames if necessary in order to create smooth transitions between motions.
In addition, the model also predicts the root trajectory.

\paragraph*{Motion Infilling}
\begin{figure}[t]
\centering
\includegraphics[width=0.48\textwidth]{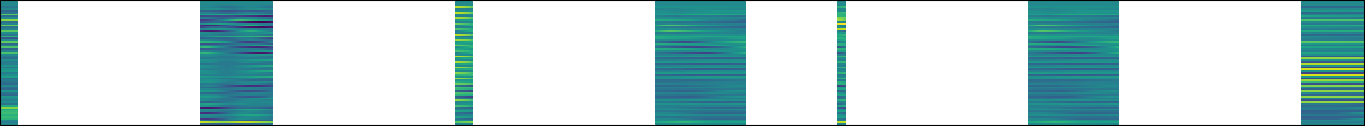}
\caption{\label{figs:inbetweening_input} Inputs for the motion infilling task. White are gaps to fill, while the rest represents the known key sequences. Please note that only a portion of the true input is shown here.}
\end{figure}

To show our model's performance in synthesizing coherent transitions between different actions we have produced a variety of results that blend several motion clips (see~\figref{fig:teaser} and more in the appendix).
To this end, we have selected a few poses from various motion sequences (called \textit{seed sequences} in the following) and chose a variable length of frames to be filled in by the model between each seed.

In ~\figref{fig:teaser} and the video we show the results of our model on blending $16$ seed sequences from a test database of different motion categories.
All seed sequences are taken from the holdout validation set.
Since our model correctly handles variable-length sequences, we place variable gaps of sizes $50$--$100$ between seed sequences.
The necessary length of a seed sequence depends on the motion category.
Experimentally we have obtained good results with ranges from $5$ (one-legged jump) to $90$ (sitting).
Typically the fewer examples of an activity are represented in the training data, the more key frames are necessary in order to constrain the model sufficiently.
\figref{fig:teaser} contains a total of $1927$ frames, of which $1300$ (roughly $67\%$) were filled in by our model.
\figref{figs:inbetweening_input} shows an excerpt of the ``image'' that is fed to the model.

The missing frames are generated efficiently by performing one forward-pass through the network. It takes $2.8$ seconds to generate $1927$ frames on a low-end GPU (i.e, NVIDIA GeForce GT 730, 2 GB memory), or $1.45$ ms per frame.
Higher end hardware (i.e., NVIDIA Quadro M6000, 12 GB memory), further reduces the inference time to a total of $0.58$ seconds ($0.3$ ms per frame).

\paragraph*{Tertiary Motion Blending}

\begin{wrapfigure}{R}{0.23\textwidth}
\centering
\vspace{-0.25cm}
\includegraphics[width=0.23\textwidth]{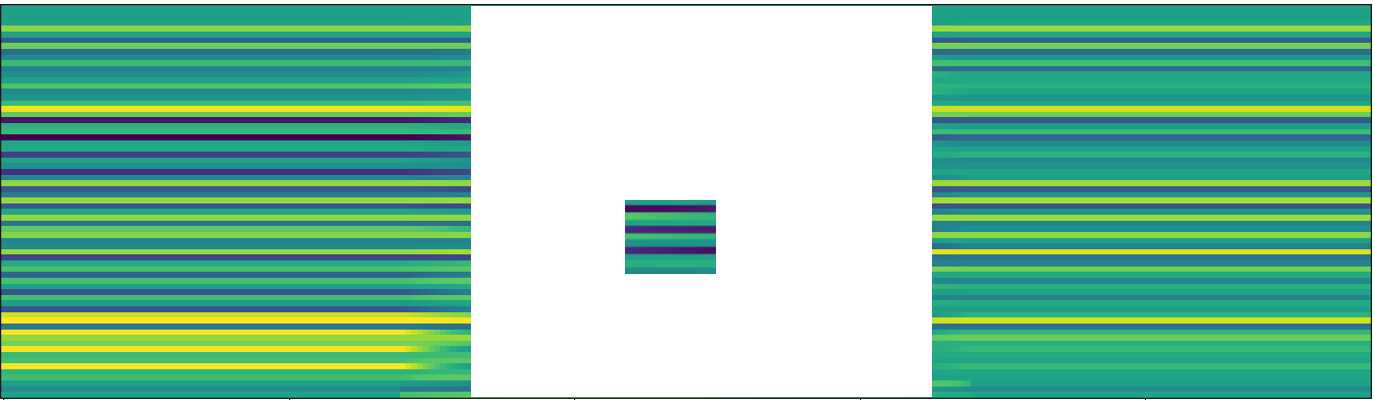}
\vspace{-0.75cm}
\end{wrapfigure}

\begin{figure}[t]
\vspace{8pt}
\includegraphics[width=\columnwidth]{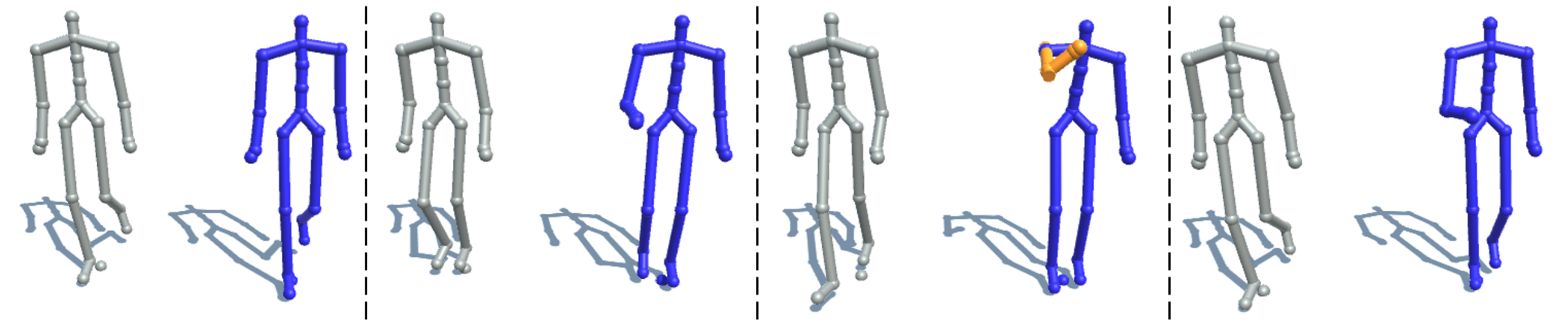}
\caption{Tertiary motion blending. Time flows from left to right. Blue: generated sequence. Orange: specific joints positions obtained from a third sequence used to alter the original motion.}
\label{figs:end_effector}
\end{figure}

\begin{figure}[t]
\vspace{8pt}
\includegraphics[width=\columnwidth]{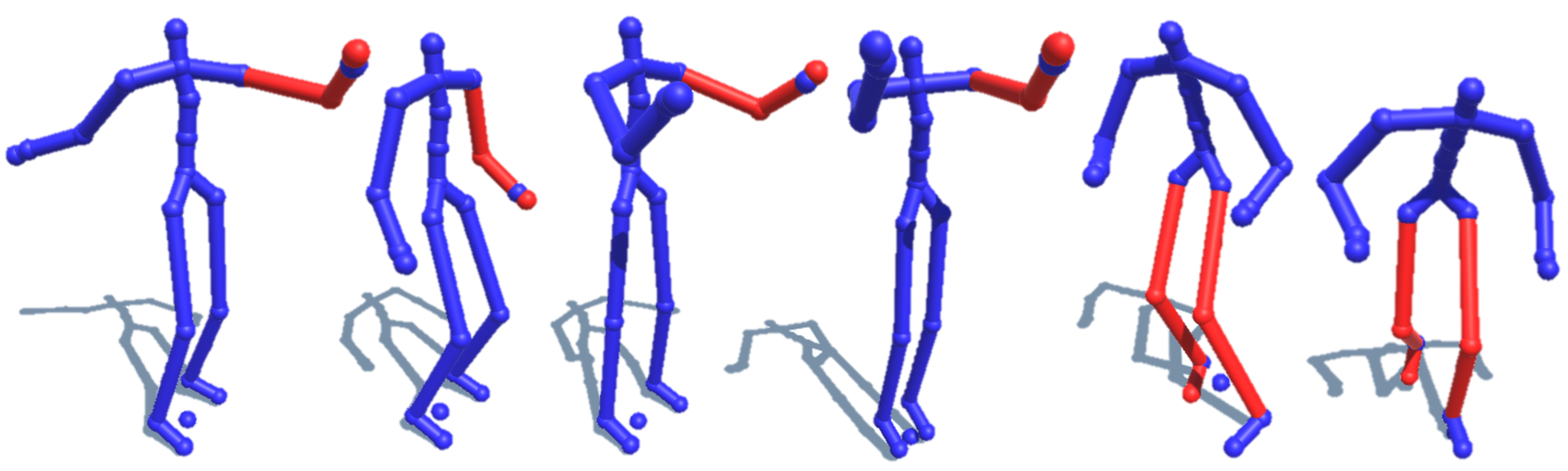}
\caption{Example of joint recovery in two sequences of $240$ frames each. In the left four characters, the left elbow and left hand (marked in red) were masked in the input. On the two right most characters the right knee, right toe and left knee were removed respectively. Outputs at different time steps are shown.}
\label{figs:joints_recovery}
\end{figure}

In addition to interpolating between key frames, our model can be used to blend in additional motion, e.g., to re-target an existing motion while constraining an end-effector.
Here we blend a grabbing action into a plain walking clip. We introduce a gap of $60$ frames in the middle of a walking sequence from the test dataset, and then place joint positions of a right arm over the length of $15$ frames into the gap (inputs shown in inset, outputs in \figref{figs:end_effector}).
The joint positions for the arm are taken from another test clip and kept in its original representation, i.e., the local body coordinates.
Note that it is necessary to provide such tertiary constraints over a number of frames, otherwise the model will treat the input as noise.

\paragraph*{Recovering Joints}

\begin{wrapfigure}{R}{0.23\textwidth}
\centering
\vspace{-0.25cm}
\includegraphics[width=0.23\textwidth]{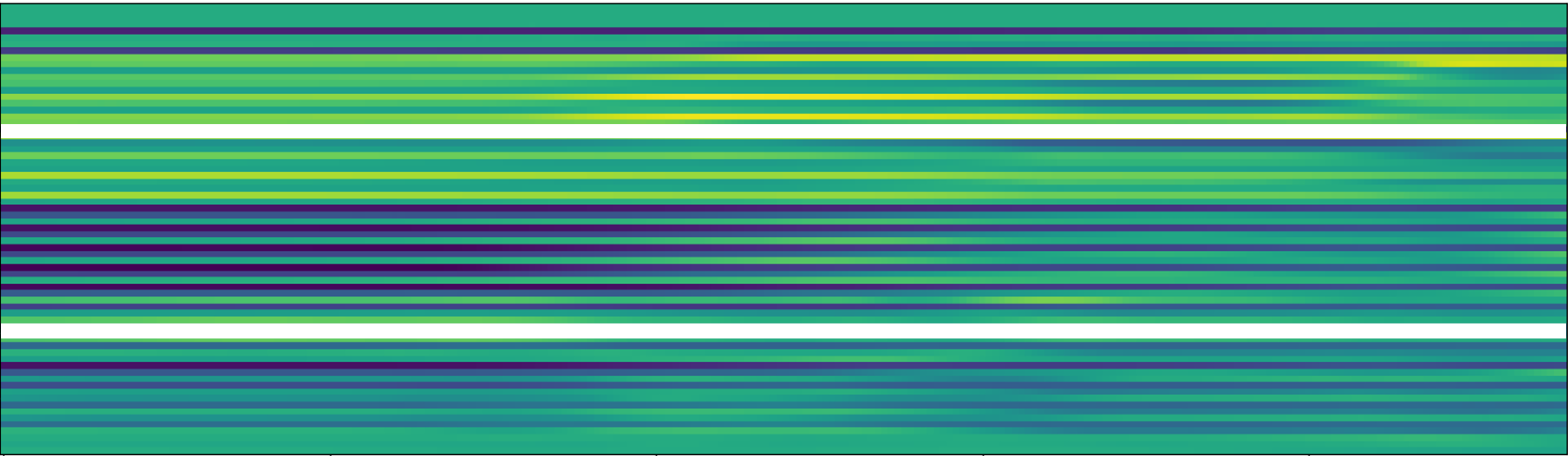}
\vspace{-0.75cm}
\end{wrapfigure}

In a further experiment we evaluate the task of recovering one or more joints that are missing over the entire sequence. This corresponds to the scenario when a tracker (e.g., an optical marker) is lost during a recording session or certain joints are occluded.
We mask different joints in various clips of length $240$ from our test dataset by removing entire rows, thus removing joints entirely over the duration of the motion sequence (see inset). \figref{figs:joints_recovery} and the accompanying video visualize the results. Despite significant loss of information, the model produces convincing looking motions, even if the missing joints are close in the hierarchy of the skeleton (e.g. elbow and hand).

\paragraph*{Variable Gap Size and Amount of Context}
A key feature of our approach is the capability to generate motion in between seed sequences with variable spacing. Here we evaluate the limits in terms of gap parameters. First, we take two seed motions and increase the gap size between the seeds incrementally. Second, instead of varying gap sizes we vary the number of given frames in the seed sequences.

For both experiments we choose walking and punching motions as seed sequences. Gap sizes 5, 20, 80, and 250 are applied with fixed seed sequences. Note that 5 and 250 are extreme values, with the latter being roughly twice the gap size seen during training. The model produces smooth and plausible motion for gap sizes 20 and 80. With gap size 250 the motion remains smooth, however towards the middle of the sequence, the model starts to converge into a mean pose before starting to prepare for the ensuing punch (see video, minute 02:47). Likewise extremely small gaps of only 5 frames in combination with drastically different source and target motions produces (overly) smooth motion and intermediate fine details are lost.

In terms of number of frames required per seed sequence we experimented with values of 1, 5, 10, 25, and 50  on either side of the gap and fixed the gap size to $80$ in line with the previous experiment. Again the model performs well under most of the configurations but the most extreme: with only a single seed frame on either end the model produces overly smooth motions and simply repositions the end-effectors. Not surprisingly, the more context we introduce, the better the predicted motion. In this particular case there was no more discernible difference in terms of motion details after 25 seed frames. Both experiments point to the same effect that not absolute numbers count, but the ratio between gap size and context provided by the seed sequences. We found this ratio should be approx.~$1/6$ for simple walking motions, $1/3$ for transitions between different motion types and $2/3$ for those containing rare activities such as ballet.

\paragraph*{Further Results}
We show results from the de-noising experiments in the appendix and the video.
Also, please refer to the appendix for more infilling results between various motion types, including types that are rare in the data.

\paragraph*{Limitations}
While we see good pose predictions across a wide range of motions and task variations there are certainly failure cases and limitations. It is possible to  successfully recover full motion cycles even if they are masked from start to end, such as a gait cycle, but only if the motion is recurring. For example, motions that are more complex and only occur once such as a full turn in-air that is masked entirely, will not be reconstructed since there is no information that would constrain the model to do so. Similarly, the reconstruction of motions that are very rare in the training data are prone to appear jittery.


\section{Conclusion}
\label{sec:conclusion}
In this paper we have proposed a deep convolutional autoencoder that learns to fill in large gaps in 3D human motion data.
The method is capable of creating smooth transitions between drastically different motion sequences and generates plausible looking motion overall.
The key idea in this approach is to cast the infilling task as an inpainting problem.
We train the neural network on image-like representations of human poses where each column represents one time step in the sequence.
During training we mask increasingly large blocks of the input data so that the model is forced to learn how to generate plausible pose data to fill in the gaps.
A curriculum learning scheme is employed to enable prediction over variable gap length and to achieve robustness against various forms of noise in the inputs.
We have evaluated our method in a number of experiments to illustrate its capabilities but also to identify its limits.

While the method produces plausible and natural motion without any further smoothing there are various areas for future work.
In particular, we currently predict complete motion sequences including relative translational and rotational root velocities.
Integrating these velocities over time results in a global root trajectory.
Clearly for the method to be applicable in a production settings it would be necessary to provide control over the global trajectories.
In the future, it would be interesting to decouple the root trajectory control from the pose generation process.

Similarly, although in our visualizations this does not seem to majorly degrade visual quality, our method cannot guarantee that bone lengths are always consistent or no foot skating artifacts occur.
Future work could look into robustifying the method in this regard, especially so in the context of multi-person datasets.
        
        \paragraph*{Acknowledgments}
        We thank Janick Cardinale for his support.
        This project has received funding from the European Research Council (ERC) under the European Union's Horizon 2020 research and innovation programme grant agreement No 717054.
        
        \begin{figure}[h]
        \vspace{-0.25cm}
        \centering
            \includegraphics[width=0.25\textwidth, trim=20 150 20 150, clip]{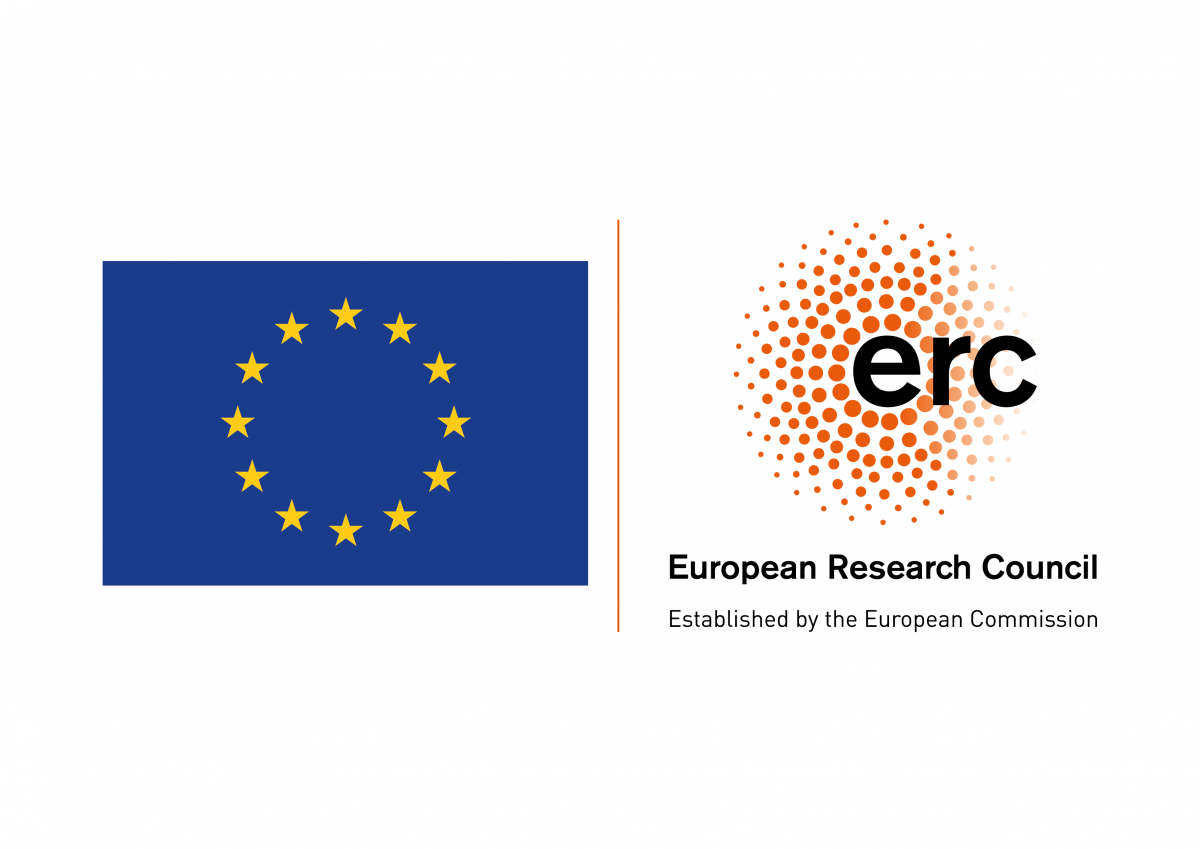}
        \end{figure}
        
        \clearpage
        {\small
        \bibliographystyle{ieee}
        \bibliography{motion-infilling}
        }

    \ifincludeappendix
    
        \clearpage
        \section{Appendix}

    \fi
\fi

\end{document}